\documentclass[letterpaper, 10 pt, conference]{ieeeconf}

\IEEEoverridecommandlockouts                              
\usepackage{makeidx}         
\usepackage[pdftex]{graphicx}
\usepackage{epstopdf}
\usepackage{multicol}        

\usepackage{hyperref}
\usepackage{amsmath,amssymb}
\usepackage{euscript}
\usepackage{subfig}
\usepackage{url}
\usepackage{mathrsfs}
\usepackage{array}
\usepackage{wrapfig}

\usepackage{epsfig,float,alltt}
\usepackage{psfrag,xr}
\usepackage{pdfpages}
\usepackage{epstopdf}
\usepackage{pstool}

\usepackage{bm}

\usepackage[T1]{fontenc}

\usepackage{multirow}

\usepackage{balance} 

\newcommand{\figref}[1]{Fig.~\ref{#1}}
\newcommand{\secref}[1]{Sec.~\ref{#1}}

\renewcommand{\epsilon}{\varepsilon}

\newcommand{\pos}{\bm{p}}

\usepackage{multirow}
\newcommand{\bom}    {\mbox{\boldmath $\omega$}}

{\end{tabbing} \end{mdframed} \vspace{5px}}

\newcommand{\BM}{\begin{bmatrix}}
\newcommand{\EM}{\end{bmatrix}}

\newcommand{\beq}{\begin{equation}}
\newcommand{\eeq}{\end{equation} }

\begin{document}

\title{Adaptive Force-based Control for Legged Robots}

\author{Mohsen Sombolestan$^{*}$, Yiyu Chen$^{*}$ and Quan Nguyen 
\thanks{This work is supported by USC Viterbi School of Engineering startup funds.}
\thanks{M. Sombolestan, Y. Chen, and Q. Nguyen are with the Department of Aerospace and Mechanical Engineering, University of Southern California, Los Angeles, CA 90089, email: {\tt somboles@usc.edu, yiyuc@usc.edu, quann@usc.edu}.}
\thanks{$^{*}$ These authors contributed equally to this work.}
}

\maketitle



\begin{abstract}

Adaptive control can address model uncertainty in control systems. 
However, it is preliminarily designed for tracking control. Recent advancements in the control of quadruped robots show that force control can effectively realize agile and robust locomotion. In this paper, we present a novel adaptive force-based control framework for legged robots. We introduce a new architecture in our proposed approach to incorporate adaptive control into quadratic programming (QP) force control. 
Since our approach is based on force control, it also retains the advantages of the baseline framework, such as robustness to uneven terrain, controllable friction constraints, or soft impacts. 
Our method is successfully validated in both simulation and hardware experiments. While the baseline QP control has shown a significant degradation in the body tracking error with a small load, our proposed adaptive force-based control can enable the 12-kg Unitree A1 robot to walk on rough terrains while carrying a heavy load of up to 6 kg (50\% of the robot weight). 
When standing with four legs, our proposed adaptive control can even allow the robot to carry up to 11 kg of load (92\% of the robot weight) with less than 5-cm tracking error in the robot height.

\end{abstract}

\section{Introduction}
\label{sec: intro}

Legged robots have great potential for applications in disaster and rescue missions. In contrast to wheeled or tracked robots, legged robots represent remarkable performance for navigating uneven terrains. 
Designing and controlling machines to realize these potentials has motivated work across the legged robotics community and highly-capable quadrupeds (e.g., \cite{7758092,Park15b, IDETC_disinfection_paper,OptimizedJump}), beginning to assist humans in demanding situations. 

Impulse-based gait design introduced in \cite{park2017high} can be used to achieve high-speed bounding for quadruped robots. Nevertheless, the approach was primarily designed for 2D motion.
The recent development of model predictive control (MPC) approach for quadruped robots \cite{di2018dynamic} utilizes convex optimization to solve for optimal ground reaction force. The approach is based on the simplified rigid body dynamics of the robot, enabling real-time computation for the controller. This framework has achieved a wide range of agile motion for 3D quadruped robots.

However, these controllers assume accurate knowledge of the dynamic model, or in other words, do not address substantial model uncertainty in the system. 
Many safety-critical missions, such as firefighting, disaster response, exploration, etc., require the robot to operate swiftly and stably while dealing with high levels of uncertainty and large external disturbances. The demand for practical requirements motivates our research on adaptive control for quadruped robots.


The introduction of the $L_1$ adaptive control technique has enabled the decoupling of adaptation and robustness in adaptive control techniques. In particular, applying a low-pass filter as part of the adaptation laws helps the $L_1$ adaptive controller to guarantee not only stability \cite{L1:StabilityMargins:ACC07} and transient performance \cite{L1:TransientPerformance:TAC08} but also smooth control inputs, which can be critical for robotics applications. Our prior work on $L_1$ adaptive control for bipedal robots \cite{nguyen20151} uses a control Lyapunov function (CLF) based controller to create a closed-loop nonlinear reference model for the $L_1$ adaptive controller. 
However, the control paradigm in this prior work is based on Hybrid Zero Dynamics \cite{GRCHSH08}, which uses joint position control to track the desired trajectory from optimization for each robot joint. 

\begin{figure}[t!]
	\center
	\includegraphics[width=0.75\linewidth]{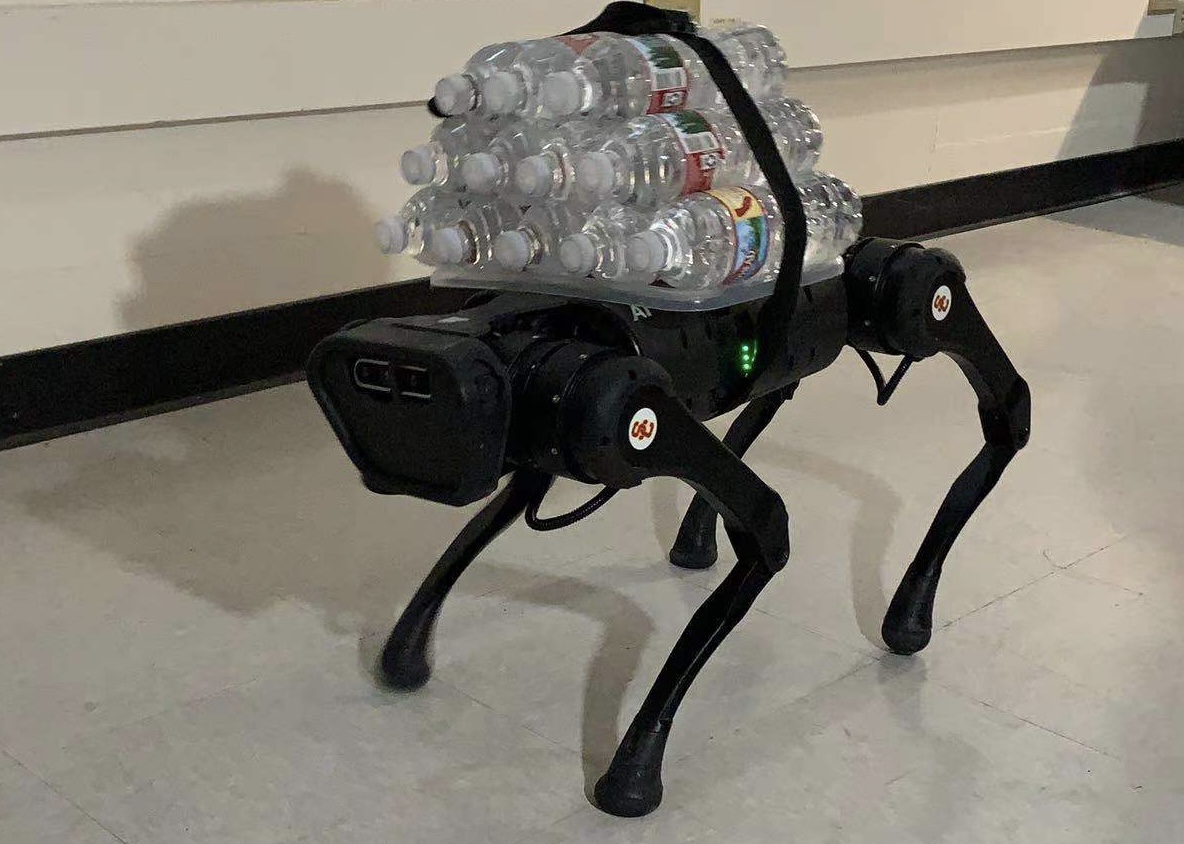}
	\caption{Our proposed adaptive force-control approach is successfully validated in experiment on the 12-kg A1 robot walking while carrying an unknown load of up to 6 kg (50\% of body weight). Experiment video: \protect\url{https://youtu.be/tWlO7b-EhP4}} 
	\label{fig:uneven terrain}
	\vspace{-1.5em}
\end{figure}

In this work, we present a novel adaptive force-based control framework to utilize the advantages of force-based control \cite{Focchi17} in dynamic legged robots including robustness to rough terrains, flexibility in obtaining a wide range of gaits, and soft impacts during locomotion \cite{bledt2018cheetah}.
Due to fundamental differences between trajectory tracking and force-based control, it requires the development of a new control architecture to integrate adaptive control into the force-based control framework.
To the best of our knowledge, this is the first adaptive force-based controller successfully developed for quadruped robots. Our approach is successfully validated in both a high-fidelity simulation and hardware experiment. Although the baseline controller fails to maintain the robot balance under small model uncertainty, our proposed adaptive controller can satisfy the expectation while carrying an unknown load up to 50\% of the robot weight (shown in \figref{fig:uneven terrain}). Thanks to the combination with the force-based controller, our approach can also allow the robot to navigate rough terrains while carrying an unknown and time-varying load.

The followings are the main contribution of the paper:
\begin{itemize}
	
	\item We introduce a novel control architecture to incorporate adaptive control into the force-based control framework to adapt to significant model uncertainty of the system dynamics.
	
	\item Since our approach is based on force control, it retains critical advantages of the baseline framework, including soft impact, robustness to rough terrains, controllable friction constraints, and the flexibility in adapting to different locomotion gaits.
	
	\item We prove that our approach yields Input-to-State (ISS) stability for the control system.
	
	
	\item We successfully validate our approach in the simulation of a quadruped robot walking on an uneven, steep slope while carrying an unknown heavy load up to 50\% of the robot weight and subjecting to unknown force disturbance to different parts of the body during walking.  
	
	\item We successfully validate our approach on the real robot hardware of A1, a 12-kg quadruped robot. With our framework, the robot stands up and balances with unknown loads of up to 11 kg, which is 92\% of the robot's weight. For comparison, the baseline non-adaptive controller can not even stand up with only 6 kg of load.
	
    \item We successfully validate our approach on the real robot hardware of the A1 robot walking stably while carrying unknown loads of up to 6 kg, which is 50\% of the robot weight. For comparison, the baseline non-adaptive controller fails to control the robot to walk with only 3 kg of load.
\end{itemize}


The remainder of the paper is organized as follows. 
\secref{sec:control_arch} presents the background on the force-based control architecture for quadruped robots. The proposed adaptive controller to compensate uncertainties is elaborated in \secref{sec:adaptive control}. Then, the stability proof of the whole system is described in \secref{sec:stability proof}. Furthermore, the numerical and experimental validation are shown in \secref{sec:simulation} and \secref{sec:experiment}, respectively. Finally, \secref{sec:conclusion} provides concluding remarks.  
\section{BACKGROUND}
\label{sec:control_arch}
\begin{figure}[bt!]
	\centering
	\includegraphics[width=1\linewidth]{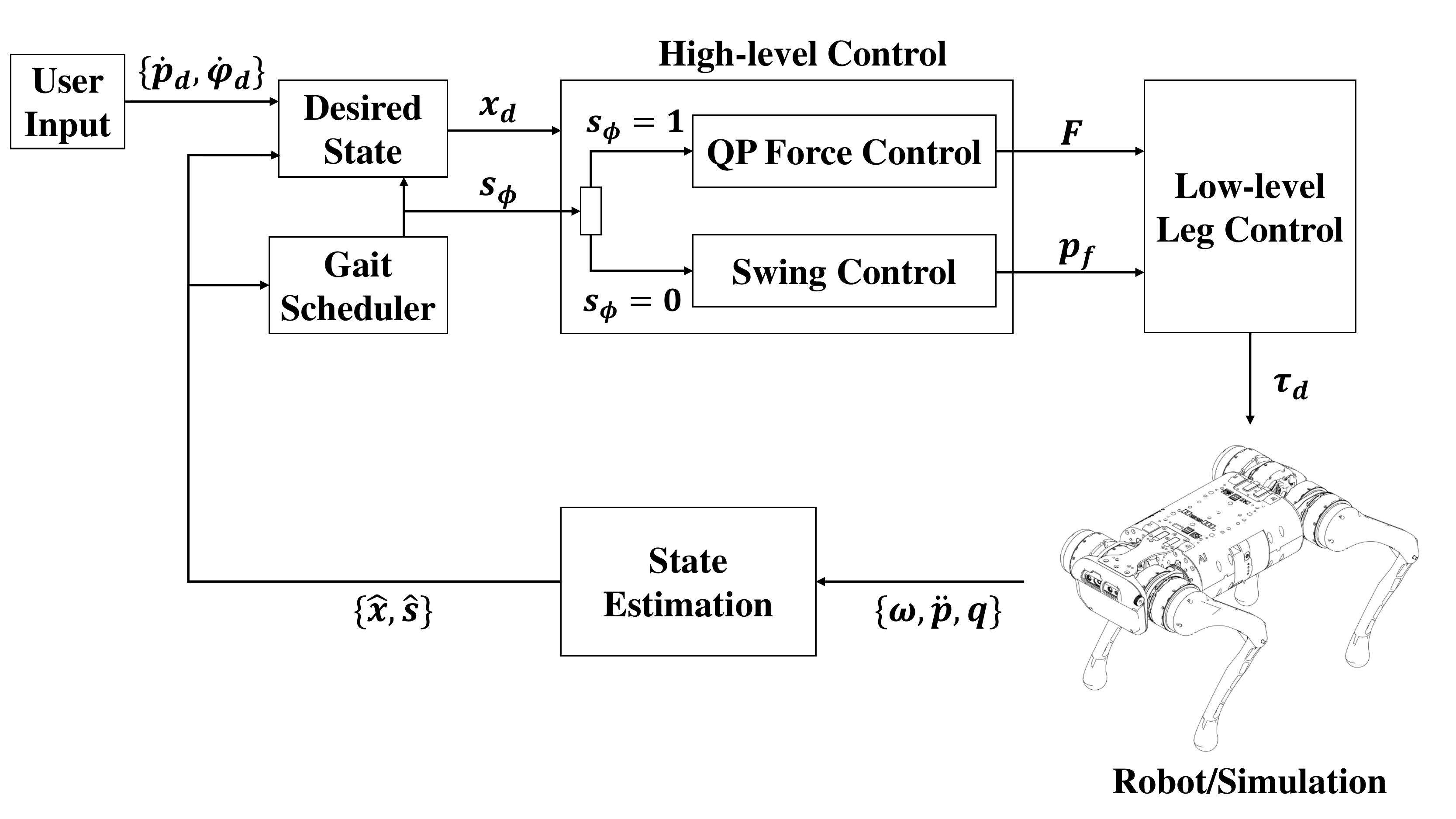}
	\caption{{\bfseries Control Architecture Overview.} Block diagram of control architecture for A1 robot.}
	\label{fig:ControlOverview}
\end{figure}

In this section, we present the background on force-based control of quadruped robots. The control architecture of the robot consists of several modules, \cite{bledt2018cheetah} including high-level controller, low-level controller, state estimation, and gait scheduler as presented in \figref{fig:ControlOverview}. From user input and state estimation, a reference trajectory can be generated for high-level control. The gait scheduler defines the gait timing and sequence to switch between swing and stance phases for each leg. 
The high-level controller calculates position control for swing legs and force control for stance legs based on the user commands and gait timing. 
The low-level leg control converts the command generated by high-level control into joint torques for each motor. Each module of the control architecture will be elaborated in the following sections. The $L_1$ adaptive controller is built on this baseline architecture and will be elaborated in \secref{sec:adaptive control}.

\subsection{Robot Specification}
\label{sec:RobotModel}
\begin{figure}[bt!]
	\center
	\includegraphics[width=0.8\linewidth]{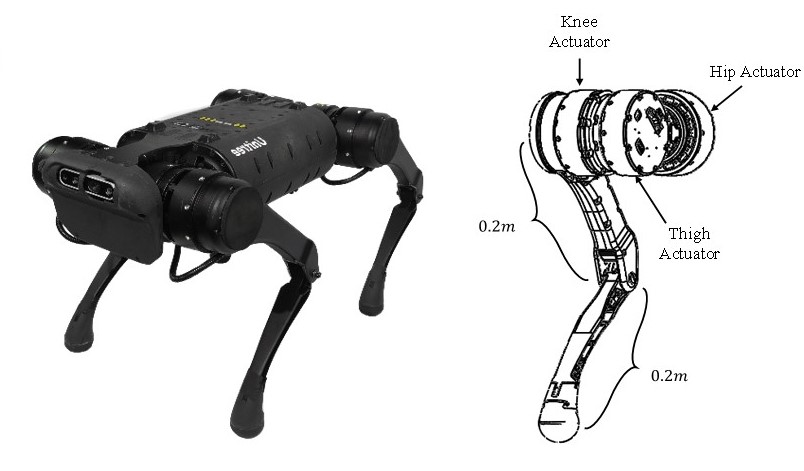}
	\caption{{\bfseries Robot Configuration.}  Overview of A1 robot and leg configuration}
	\label{fig:RobotDiagram}
	\vspace{-1.5em}
\end{figure}

In this paper, we will validate our controller on Unitree A1, a mini dynamic quadruped robot (see \figref{fig:RobotDiagram}). The A1 robot is weighted 12 kg and has low-inertial legs. The robot is equipped with high torque density electric motors using planetary gear reduction. It is capable of ground force control without using any force or torque sensors. The A1 robot uses these high-performance actuators for all the hip, thigh, and knee joints to enable full 3D control of ground reaction forces. It is also equipped with contact sensors on each foot which are used for contact detection.

Each of A1's actuators consists of a custom high torque density electric motor coupled to a single-stage 9:1 planetary gear reduction. The lower link is driven by a bar linkage that passes through the upper link. The legs are serially actuated, but to keep leg inertia low, the hip and knee actuators are co-axially located at the hip of each leg. Each robot joint has the maximum torque of $33.5~(Nm)$ and the maximum speed of $21~(rad/s)$.


\subsection{Gait Scheduler}
The A1's gait is defined by a finite state machine using a leg-independent phase variable to schedule contact and swing phases for each leg \cite{bledt2018cheetah}. The gait scheduler utilizes independent boolean variables to define contact states scheduled $\bm{s}_{\phi} \in \{1 = contact, 0 = swing\}$ and switch each leg between swing and stance phases. Based on the contact schedule, the controller will execute either position control during swing or force control during stance for each leg.
In this paper, in order to introduce significant model uncertainty to the system dynamics, we focus on the application of load-carrying task, where the load is unknown to the robot or the control system. 
Having more legs on the ground during walking could also mean that the robot could produce a larger total ground reaction force to support the heavy load. Therefore, for this task, we use quasi-static walking gait to maximize the number of legs on the grounds during walking (i.e., 3 stance legs and 1 swing leg throughout the gait). Note that while we decide to use this gait to better show the effectiveness of our approach in addressing model uncertainty, our framework is not limited by any specific gait. Similar to the baseline force-based control approach, the approach can work for different gaits by only changing the gait definition in the gait scheduler. 

\subsection{Simplified Dynamics for Control Design}
Due to the inherent nonlinear nature of the legged system, we utilized a simplified rigid-body dynamics to optimize the ground reaction forces to balance the whole body motion and enable real-time optimization. By design, the robot has light limbs with low inertia as compared to the overall body. Therefore, it is reasonable to ignore the effects of the legs on the whole body motion to plan ground reaction forces. In particular, the A1 robot controller model employs a commonly used linear relationship \cite{Focchi17,StephensAtkeson10b} between the body’s linear acceleration ($\ddot{\pos}_{c}\in \mathbb{R}^{3}$), angular acceleration ($\dot \bom_b\in \mathbb{R}^{3}$), and the ground reaction forces ($\bm{F} = [\bm{F}_1^T, \bm{F}_2^T, \bm{F}_3^T, \bm{F}_4^T]^T \in \mathbb{R}^{12}$) acting on each of the robot’s four feet. The following linear model is expressed:
\begin{align}
\label{eq:linear_model}
\underbrace{\left[\begin{array}{ccc} \mathbf{I}_{3} & \dots & \mathbf{I}_{3}  \\ \hspace{-0.1cm}[\bm{p}_{1} - \pos_{c}] \times & \dots & [\bm{p}_{4} - \pos_{c}] \times \end{array} \right]}_{\bm{A}\in \mathbb{R}^{6 \times 12}} \bm{F} = \underbrace{\left[\begin{array}{c} m (\ddot{\pos}_{c} +\bm{g}) \\ \bm{I}_G \dot \bom_b \end{array} \right]}_{\bm{b} \in \mathbb{R}^{6}},
\end{align}
where $m$ and $\bm{I}_G\in \mathbb{R}^{3 \times 3}$ are the robot's mass and moment of inertia
, $\bm{g}\in \mathbb{R}^{3}$ is the gravity vector, $\bm{p}_{c}\in \mathbb{R}^{3}$ is the position of the center of mass (COM), and $\bm{p}_{i}\in \mathbb{R}^{3}$ ($i \in \{1,2,3,4\}$) are the positions of the feet. The term $[\bm{p}_{i} - \pos_{c}] \times$ is the skew-symmetric matrix representing the cross product $(\bm{p}_{i} - \pos_{c}) \times  \bm{F}_i$. The term $\bm{I}_G \dot \bom_b$ is actually an approximation of following equation:
\begin{align}
\frac{d}{dt}(\bm{I}_G \bom_b) = \bm{I}_G \dot \bom_b + \bom_b \times (\bm{I}_G \bom_b) \approx \bm{I}_G \dot \bom_b.
\end{align}
Under the assumption of small angular velocities, the term $\bom_b \times (\bm{I}_G \bom_b)$ is relatively small and therefore will be ignored in this framework (see \cite{Focchi17}).
The vector $\bm{b}$ in \eqref{eq:linear_model} can be rewritten as:
\begin{align}\label{eq:b}
\bm{b} = \underbrace{\left[\begin{array}{cc} m\mathbf{I}_3 & \bm{0}_3 \\ \bm{0}_3 & \bm{I}_G \end{array} \right]}_{\bm{M}\in \mathbb{R}^{6 \times 6}} \left[\begin{array}{c}  \ddot{\pos}_{c} \\  \dot \bom_{b} \end{array} \right] + \underbrace{\left[\begin{array}{c}  m \bm{g} \\  \bm{0} \end{array} \right]}_{\bm{G}\in \mathbb{R}^{6}}.
\end{align}

\subsection{Balance Controller}
Since the model \eqref{eq:linear_model} is linear, the controller can naturally be formulated as the following quadratic programming (QP) problem \cite{Gehring13}, which can be solved in real-time at $1~kHz$: 
\begin{align}
\nonumber
\bm{F}^* =  \underset{\bm{F} \in \mathbb{R}^{12}}{\operatorname{argmin}}   \:\:  &(\bm{A} \bm{F} - \bm{b}_d)^T \bm{S} (\bm{A} \bm{F} - \bm{b}_d)  \\ \label{eq:BalanceControlQP} & + \gamma_1 \| \bm{F} \|^2 + \gamma_2 \| \bm{F} - \bm{F}_{\textrm{prev}}^* \|^2\\
\mbox{s.t. }& \quad \:\: \underline{\bm{d}} \leq \bm{C} \bm{F} \leq \bar{\bm{d}} \nonumber\\
& \quad \:\: \bm{F}_{swing}^z=0 \nonumber
\end{align}
where $\bm{b}_d$ is the desired dynamics and will be described in \secref{sec:adaptive control}. 
The cost function in \eqref{eq:BalanceControlQP} includes terms that consider three goals, including (1) driving the COM position and orientation to the desired trajectories; (2) minimizing the force commands; and (3) filtering the change of the current solution $\bm{F}^*$ with respect to the solution from the previous time-step, $\bm{F}^*_{prev}$. 
The priority of each goal in the cost function is defined by the weight parameters $\bm{S}\in \mathbb{R}^{6 \times 6}$, $\gamma_1$, $\gamma_2$ respectively.
The constraints in the QP formulation enforce friction constraints, input saturation, and contact constraints.
More details about this QP controller can be seen in \cite{Focchi17}.

Besides the friction constraint, we will enforce the force constraints for the swing legs, $\bm{F}_{swing}=\bm{0}$. The swing legs are then kept at the posing position using PD control described \cite{bledt2018cheetah} until it switches to stance phase. Based on this controller, a walking controller with a static walking gait is implemented on the robot. 

\section{PROPOSED APPROACH: ADAPTIVE FORCE-BASED CONTROL}
\label{sec:adaptive control}
Based on the control architecture in \secref{sec:control_arch}, in this Section, we will present a novel control architecture to integrate adaptive control into the force control framework. While our approach is not limited to any specific adaptive control approach, we decide to use $L_1$ adaptive control \cite{L1:book:FastAdaptation,nguyen20151} thanks to its advancement in guaranteeing fast adaptation and smooth control signals. 

Our prior work on \cite{nguyen20151} introduced adaptive control for bipedal robots based on Hybrid Zero Dynamics (HZD) \cite{WEGRKO03}, which uses joint position control to track the desired trajectory designed by trajectory optimization. HZD is commonly used for control of bipedal robots to address hybrid and underactuated dynamics of the system. The approach was successfully validated for walking \cite{nguyen2018dynamic} and running \cite{SrPaPoGr2013} on dynamic bipedal robots. 
In this paper, our approach is, however, based on force control, which optimizes ground reaction forces (GRFs) to achieve dynamic locomotion for legged robots \cite{bledt2018cheetah}. 
The force control approach is robust to rough terrains \cite{Focchi17} and capable of realizing a wide range of dynamic motions \cite{di2018dynamic}. Therefore, in addition to adapting to model uncertainty, utilizing force control could maintain these advantages for the system.
In this section, we present a proposed control architecture to incorporate adaptive control into the force-based control framework.


\subsection{Closed-loop Dynamics}
To incorporate adaptive control into the force-based control framework, we firstly reformulate the system as follow.

Let the state variable define as $\bm{\eta} = [\bm{e},~\dot{\bm{e}}]^T \in \mathbb{R}^{12}$, with
\begin{align}
\bm{e} = \left[\begin{array}{c} \bm{p}_{c}-\bm{p}_{c,d} \\ \log(\bm{R}_d \bm{R}^T)	\end{array} \right]\in \mathbb{R}^{6}, \quad 
\dot{\bm{e}} = \left[\begin{array}{c} \dot{\bm{p}}_{c}-\dot{\bm{p}}_{c,d} \\ \bom_b -\bom_{b,d} \end{array} \right]\in \mathbb{R}^{6},
\end{align} 
where $\bm{p}_{c,d}\in \mathbb{R}^{3}$ is the desired position of COM, $\dot{\bm{p}}_{c,d}\in \mathbb{R}^{3}$ is the desired body's linear velocity, and $\bom_{b,d}\in \mathbb{R}^{3}$ is the desired body's angular velocity. The desired and actual body orientations are described using rotation matrices $\bm{R}_d\in \mathbb{R}^{3 \times 3}$ and $\bm{R}\in \mathbb{R}^{3 \times 3}$, respectively. The orientation error is obtained using the exponential map representation of rotations \cite{Bullo95, MurrayLiSastry94}, where the $log(.):\mathbb{R}^{3 \times 3} \to \mathbb{R}^{3} $ is a mapping from a rotation matrix to the associated rotation vector \cite{Focchi17}.
Therefore, the closed-loop error dynamics in state-space form can be represented as follow:
\begin{align}
\label{fgBarDynamics}
\dot{\bm{\eta}} = \bm{D} \bm{\eta} + \bm{H} \bm{u},
\end{align}
where
\begin{align}
\bm{D} = \left[ \begin{array}{cc} \bm{0}_6  & \bm{1}_6 \\ \bm{0}_6 & \bm{0}_6 \end{array} \right]\in \mathbb{R}^{12 \times 12}, \quad
\bm{H} =  \left[ \begin{array}{c} \bm{0}_6 \\ \bm{1}_6 \end{array} \right] \in \mathbb{R}^{12 \times 6},
\end{align}
and $\bm{u}\in \mathbb{R}^{6}$ is the control input function. By employing a PD control law, we have
\begin{equation}
\bm{u} = \begin{bmatrix}-\bm{K}_P & -\bm{K}_D\end{bmatrix} \bm{\eta},
\label{PDcontrol}
\end{equation}
where $\bm{K}_P \in \mathbb{R}^{6 \times 6}$ and  $\bm{K}_D \in \mathbb{R}^{6 \times 6}$ are diagonal positive definite matrices.

The goal of the controller is to find out optimal leg forces $\bm{F}$ that achieve the control input function described above and accordingly maintain the error (state variable $\bm{\eta}$) within a bounded range. Thus, we need to find a relation between the linear model \eqref{eq:linear_model} and the closed-loop error dynamics \eqref{fgBarDynamics}.

First, from equation \eqref{fgBarDynamics} it can be obtained that
\begin{align}\label{eq:epp}
\ddot{\bm{e}} = \left[\begin{array}{c} \ddot{\pos}_{c} - \ddot{\pos}_{c,d} \\ \dot \bom_{b} - \dot \bom_{b,d} \end{array} \right] = \bm{u},
\end{align}
where $\ddot{\pos}_{c,d}$ and $\dot \bom_{b,d}$ are the desired COM linear acceleration and the desired angular acceleration, respectively. The desired trajectory for the robot is obtained from the velocity command. Therefore, both $\ddot{\pos}_{c,d}$ and $\dot \bom_{b,d}$ are zero vectors. Then from \eqref{eq:b} and \eqref{eq:epp}, the desired dynamics can be given by
\begin{align}\label{eq:b_d}
\bm{b}_d = \bm{M} \bm{u} + \bm{G}, 
\end{align}
where $M$ and $G$ are defined in \eqref{eq:b}.
\subsection{Effects of uncertainty in dynamic}
The QP formulation described in \eqref{eq:BalanceControlQP} provides input-to-state stability for quadruped during walking and standing that requires the accurate dynamical model of the system. The uncertainty comes from the mass, inertia, or rough terrain that has adverse effects on the dynamics of the system. Sometimes, it may cause instability in the control of the robot.   	

If we consider uncertainty in the dynamics and assume that the matrices $\bm{M}$ and $\bm{G}$ of the real dynamics are unknown, we then have to design our controller based on nominal matrices $\bar{\bm{M}}$, $\bar{\bm{G}}$. Therefore, the desired dynamic can be represented as
\begin{align}
\bm{b}_d = \bar{\bm{M}} (\bm{u} + \bm{\theta}) + \bar{\bm{G}}
\end{align}
where,
\begin{align}
\bm{\theta} = \bar{\bm{M}}^{-1} \left[(\bm{M}-\bar{\bm{M}})\bm{u} + (\bm{G}-\bar{\bm{G}}) \right]\in \mathbb{R}^{6},
\end{align}
and the closed-loop system now takes the form
\begin{equation}
\label{EtaClosedLoopUncertainty}
\dot{\bm{\eta}}=\bm{D}\bm{\eta}+\bm{H}(\bm{u}+\bm{\theta}).
\end{equation}

\subsection{$L_1$ adaptive controller for compensating the uncertainties}
From the previous subsection, we describe the system with uncertainty by \eqref{EtaClosedLoopUncertainty}. As a result, for any time $t$, we can always find out $\bm{\alpha}(t)\in \mathbb{R}^{6}$ and $\bm{\beta}(t)\in \mathbb{R}^{6}$ such that \cite{L1:UnkownNonlinearities:ACC08}:
\begin{align}
\bm{\theta}(\bm{\eta},t)=\bm{\alpha}(t)||\bm{\eta}||+\bm{\beta}(t).
\end{align}

\begin{figure}[bt!]
	\center
	\includegraphics[width=1.0\linewidth]{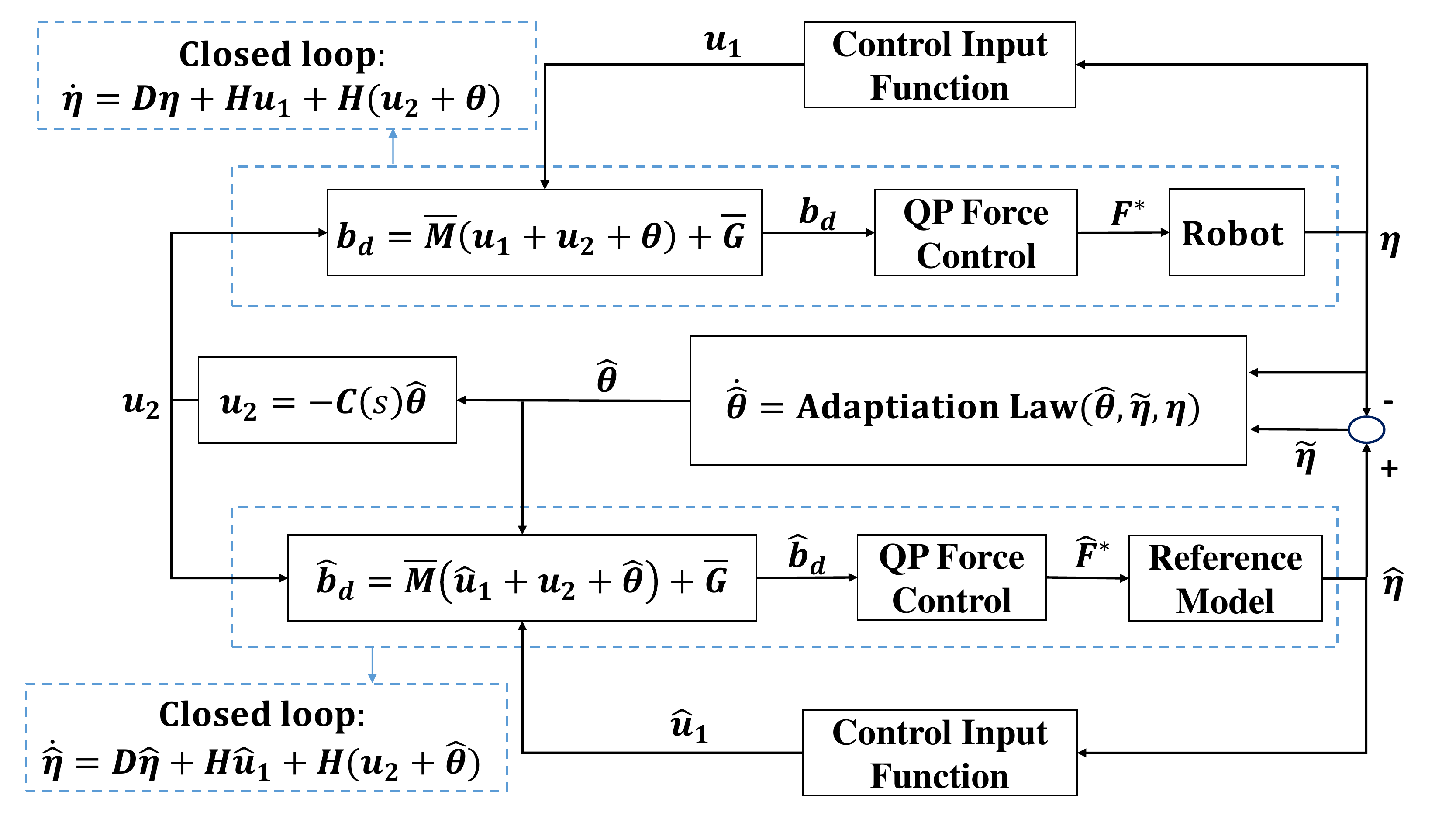}
	\caption{Block diagram of the proposed adaptive force-based controller.
	}
	\label{fig:ControlDiagram}
\end{figure}

The goal of our method is to design a combined controller $\bm{u}=\bm{u}_1+\bm{u}_2$, where $\bm{u}_1$ is to control the model to follow the desired reference model and $\bm{u}_2$ is to compensate the nonlinear uncertainties $\bm{\theta}$. The reference model is similar to linear model described in \eqref{eq:linear_model} which, instead of $\bm{M}$ and $\bm{G}$, the nominal matrices ($\bar{\bm{M}}$, $\bar{\bm{G}}$) are being used. Moreover, the model updates itself in real-time using ODE solvers. The diagram of our proposed force-based adaptive control is presented in \figref{fig:ControlDiagram}.

We present a method to consider a reference model for $L_1$ adaptive control that arises from a QP controller with input-to-state stability describe in \eqref{eq:BalanceControlQP}. The state predictor can then be expressed as follows,
\begin{align}
\label{ref_model}
\dot{\hat{\bm{\eta}}}=\bm{D}\hat{\bm{\eta}}+\bm{H}\hat{\bm{u}}_{1}+\bm{H}(\bm{u}_2+\hat{\bm{\theta}}),
\end{align}
where,
\begin{align}
\hat{\bm{\theta}}=\hat{\bm{\alpha}}||\bm{\eta}||+\hat{\bm{\beta}},
\end{align}
and $\hat{\bm{u}}_1$ is defined as:
\begin{equation}
\hat{\bm{u}}_1 = \begin{bmatrix}-\bm{K}_P & -\bm{K}_D\end{bmatrix} \hat{\bm{\eta}}. \end{equation}
According to \eqref{EtaClosedLoopUncertainty}, the $\hat{\bm{b}}_d$ get the form
\begin{align}
\hat{\bm{b}}_d = \bar{\bm{M}} (\hat{\bm{u}}_1 + \bm{u}_2 + \hat{\bm{\theta}}) + \bar{\bm{G}},
\end{align}
and the optimal distribution of leg forces $\hat{\bm{F}}$ for the reference model can be achieved by
\begin{align}
\nonumber
\hat{\bm{F}}^* =  \underset{\hat{\bm{F}} \in \mathbb{R}^{12}}{\operatorname{argmin}}   \:\:  &(\hat{\bm{A}} \hat{\bm{F}} - \hat{\bm{b}}_d)^T \bm{S} (\hat{\bm{A}} \hat{\bm{F}} - \hat{\bm{b}}_d)  \\ \label{eq:BalanceControlQP_RF} & + \gamma_1 \| \hat{\bm{F}} \|^2 + \gamma_2 \| \hat{\bm{F}} - \hat{\bm{F}}_{\textrm{prev}}^* \|^2\\
\mbox{s.t. }& \quad \:\:  \bm{C} \hat{\bm{F}} \leq \bm{d} \nonumber\\
& \quad \:\: \hat{\bm{F}}_{swing}^z=0 \nonumber.
\end{align}

In order to compensate the estimated uncertainty $\hat{\bm{\theta}}$, we can just simply choose $\bm{u}_2=-\hat{\bm{\theta}}$ to obtain
\begin{equation}
\dot{\hat{\bm{\eta}}}=\bm{D}\hat{\bm{\eta}}+\bm{H}\hat{\bm{u}}_{1}.
\end{equation}
However, $\hat{\bm{\theta}}$ typically has high frequency due to fast estimation in the adaptation law.  For the reliability and robustness of the control scheme, it is essential to obtain smooth control signals, especially for robotic applications. Thus, we apply the $L_1$ adaptive control scheme to decouple estimation and adaptation 
\cite{L1:TransientPerformance:TAC08}. Therefore, we will have
\begin{align}
\label{controller mu2}
\bm{u}_2=-C(s)\hat{\bm{\theta}},
\end{align}
where $C(s)$ is a second-order low-pass filter with a magnitude of 1: 

\begin{align}
\label{eq:Cs}
C(s) = \frac{{\omega_n}^2}{s^2 + 2 \zeta \omega_n s+ {\omega_n}^2} . 
\end{align}

Define the difference between the real model and the reference model $\tilde{\bm{\eta}}=\hat{\bm{\eta}}-\bm{\eta}$, we then have,
\begin{align}
\dot{\tilde{\bm{\eta}}}=\bm{D}\tilde{\bm{\eta}}+\bm{H}\tilde{\bm{u}}_{1}+\bm{H}(\tilde{\bm{\alpha}}||\bm{\eta}||+\tilde{\bm{\beta}}),
\end{align}
where
\begin{align}
\tilde{\bm{u}}_{1}=\hat{\bm{u}}_{1}-\bm{u}_1,~
\tilde{\bm{\alpha}}=\hat{\bm{\alpha}}-\bm{\alpha},~
\tilde{\bm{\beta}}=\hat{\bm{\beta}}-\bm{\beta}.
\end{align}

As a result, we will estimate $\bm{\theta}$ indirectly through $\bm{\alpha}$ and $\bm{\beta}$, or the values of $\hat{\bm{\alpha}}$ and $\hat{\bm{\beta}}$ computed by the following adaptation laws based on the projection operators \cite{ProjOperator},
\begin{align}
\label{adap_law}
\dot{\hat{\bm{\alpha}}}=\bm{\Gamma}\text{Proj}(\hat{\bm{\alpha}},\bm{y}_{\alpha}),~
\dot{\hat{\bm{\beta}}}=\bm{\Gamma}\text{Proj}(\hat{\bm{\beta}},\bm{y}_{\beta}).
\end{align}
where $\bm{\Gamma} \in \mathbb{R}^{6 \times 6}$ is a symmetric positive definite matrix. The projection functions $\bm{y}_{\alpha}\in \mathbb{R}^{6}$ and $\bm{y}_{\beta}\in \mathbb{R}^{6}$ are
\begin{align}\label{eq:proj_fun}
\bm{y}_{\alpha}&=-\bm{H}^T \bm{P}\tilde{\bm{\eta}}||\bm{\eta}||, \nonumber \\
\bm{y}_{\beta}&=-\bm{H}^T \bm{P}\tilde{\bm{\eta}},
\end{align}
where $\bm{P}\in \mathbb{R}^{12 \times 12}$ will be defined in \secref{sec:stability proof}.

\section{STABILITY OF PROPOSED SYSTEM}
\label{sec:stability proof}

The goal of the QP formulation described in \eqref{eq:BalanceControlQP} is to find a solution that drives the real dynamics $\bm{A} \bm{F}$ to the desired dynamics $\bm{b}_d$.
Nevertheless, if the desired dynamic vector $\bm{b}_d$ violates the inequality constraints (such as force limits and friction constraints), the controller provides the optimal solution $\bm{F}^*$ that may not satisfy the desired dynamics. 
With this solution, we define:
\begin{align}
{\bm{b}_d}^* = \bm{A} \bm{F}^*, 
\end{align}
\begin{align}
\bm{\bm{u}}^* = \bm{M}^{-1}({\bm{b}_d}^* - \bm{G}).
\end{align}
Based on the friction constraints present in \cite{Focchi17}, the value of $\bm{F}^*$ is always bounded. Besides, according to the definition of $\bm{A}$, $\bm{M}$, and $\bm{G}$, these matrices also have bounded values. Thus, it implies that
\begin{align}
\label{bounded_mu}
\|\bm{u}^* \| \leq \delta_{{u^*}}.
\end{align}

\subsection{Linear quadratic Lyapunov theory}\label{subsec:LQL}
According to Lyapunov, the theory \cite{RESCLF:AmGaGrSr:TAC12}, the PD control described in \eqref{PDcontrol} will asymptotically stabilize the system if
\begin{equation}
\bm{A}_m = \begin{bmatrix}\bm{0}_6 & \bm{I}_6 \\-\bm{K}_P & -\bm{K}_D\end{bmatrix} \in \mathbb{R}^{12 \times 12}
\label{Amatrix}
\end{equation}
is Hurwitz.	This means that by choosing a control Lyapunov function candidate as follows
\begin{equation}
\label{VUnscaledCoords}
V(\bm{\eta}) = \bm{\eta}^{T} \bm{P} \bm{\eta},
\end{equation}
where $\bm{P}\in \mathbb{R}^{12 \times 12}$ is the solution of the Lyapunov equation 
\begin{equation}\label{eq:Lyp}
{\bm{A}_m}^T \bm{P} + \bm{P} \bm{A}_m = -\bm{Q}, 
\end{equation}
and $\bm{Q}\in \mathbb{R}^{12 \times 12}$ is any symmetric positive-definite matrix. We then have
\begin{align}\label{eq:LQL property}
\dot{V}(\bm{\eta},\bm{u}) + \lambda V(\bm{\eta}) = &
\bm{\eta}^T (\bm{D}^T \bm{P} + \bm{P} \bm{D}) \bm{\eta} \nonumber\\ &+ \lambda V(\bm{\eta}) +2 \bm{\eta}^T \bm{P} \bm{H} \bm{u} \leq 0,
\end{align}
where,
\begin{align} 
\lambda = \frac{\lambda_{min}(\bm{Q})}{\lambda_{max}(\bm{P})} > 0.
\end{align}
As a result, the state variable $\bm{\eta}$ and the control input $\bm{u}$ always remain bounded.
\begin{align}
\label{bounded eta}
\|\bm{\eta} \| \leq \delta_{\eta}, \quad \|\bm{u} \| \leq \delta_{u}.
\end{align}

However, the control signal $\bm{u}^*$ we construct by solving QP problem \eqref{eq:BalanceControlQP}, is not always the same as $\bm{u}$. Therefore, it can be rewritten as
\begin{align}\label{eq:mu_star}
\bm{\Delta} = \bm{u}^* - \bm{u}
\end{align}
where $\bm{\Delta}\in \mathbb{R}^{6}$ is the difference caused by QP optimization between the desired PD control signal $\bm{u}$ and the real signal $\bm{u}^*$. This vector is also bounded according to \eqref{bounded_mu} and \eqref{bounded eta}
\begin{align}\label{bounded_delta}
\|\bm{\Delta} \| \leq \delta_{\Delta}.
\end{align}
By substituting $\bm{u}^*$ in \eqref{eq:LQL property}, we have,
\begin{equation}\label{eq:LQL_mu_star}
\dot{V}(\bm{\eta},\bm{u}^*) + \lambda V(\bm{\eta}) \leq 2 \bm{\eta}^T \bm{P} \bm{H}\bm{\Delta} \leq \epsilon_{V},
\end{equation}
where
\begin{equation}\label{eq:epsilon_v}
\epsilon_{V} = 2 \|\bm{P}\| \delta_{\eta} \delta_{\Delta}.
\end{equation}

\subsection{Stability Proof}
We consider the following control Lyapunov candidate function
\begin{align}
\label{CLF tilde}
\tilde{V}=\tilde{\bm{\eta}}^{T}\bm{P}\tilde{\bm{\eta}}+\tilde{\bm{\alpha}}^{T}\bm{\Gamma}^{-1}\tilde{\bm{\alpha}}+\tilde{\bm{\beta}}^{T}\bm{\Gamma}^{-1}\tilde{\bm{\beta}},
\end{align}
therefore, its time derivative will be
\begin{align}
\label{dotV}
\dot{\tilde{V}}=\dot{\tilde{\bm{\eta}}}^{T}\bm{P}\tilde{\bm{\eta}}+\tilde{\bm{\eta}}^{T}\bm{P}\dot{\tilde{\bm{\eta}}} +
\dot{\tilde{\bm{\alpha}}}^{T}\bm{\Gamma}^{-1}\tilde{\bm{\alpha}}+\tilde{\bm{\alpha}}^{T}\bm{\Gamma}^{-1}\dot{\tilde{\bm{\alpha}}}
\nonumber \\
+
\dot{\tilde{\bm{\beta}}}^{T}\bm{\Gamma}^{-1}\tilde{\bm{\beta}}+\tilde{\bm{\beta}}^{T}\bm{\Gamma}^{-1}\dot{\tilde{\bm{\beta}}}, 
\end{align}
in which we have
\begin{align}
\label{CLF eta parts}
\dot{\tilde{\bm{\eta}}}^{T}\bm{P}\tilde{\bm{\eta}}+\tilde{\bm{\eta}}^{T}\bm{P}\dot{\tilde{\bm{\eta}}} 
&=
(\bm{D}\tilde{\bm{\eta}}+\bm{H}{\tilde{\bm{u}}_{1}}^*)^{T}\bm{P}\tilde{\bm{\eta}}
+
\tilde{\bm{\eta}}^{T}\bm{P}(\bm{D}\tilde{\bm{\eta}}+\bm{H}{\tilde{\bm{u}}_{1}}^*) \nonumber \\
&~~+
\tilde{\bm{\alpha}}^{T}\bm{H}^{T}||\bm{\eta}||\bm{P}\tilde{\bm{\eta}}+\tilde{\bm{\eta}}^{T}\bm{P}\bm{H}\tilde{\bm{\alpha}}||\bm{\eta}|| \nonumber \\
&~~
+\tilde{\bm{\beta}}^{T}\bm{H}^{T}\bm{P}\tilde{\bm{\eta}}+\tilde{\bm{\eta}}^{T}\bm{P}\bm{H}\tilde{\bm{\beta}}. 
\end{align}

Because $\tilde{\bm{\eta}}=\hat{\bm{\eta}}-\bm{\eta}$ satisfies the condition imposed by \eqref{eq:LQL_mu_star}, it implies that
\begin{align}
\label{RES eta tilde}
(\bm{D}\tilde{\bm{\eta}}+\bm{H}{\tilde{\bm{u}}_{1}}^*)^{T}\bm{P}\tilde{\bm{\eta}}
+
\tilde{\bm{\eta}}^{T}\bm{P}(\bm{D}\tilde{\bm{\eta}}+\bm{H}{\tilde{\bm{u}}_{1}}^*)
\le
-\lambda\tilde{\bm{\eta}}^{T}\bm{P}\tilde{\bm{\eta}} + \epsilon_{\tilde{V}},
\end{align}
where
\begin{equation}\label{eq:epsilon_v_tilde}
\epsilon_{\tilde{V}} = 2 \|\bm{P}\| \delta_{\tilde{\eta}} \delta_{\tilde{\Delta}}.
\end{equation}

Furthermore, with the property of the projection operator \cite{ProjOperator}, we have:
\begin{align}
\label{proj_property}
(\hat{\bm{\alpha}}-\bm{\alpha})^{T}(\text{Proj}(\hat{\bm{\alpha}},\bm{y}_{\alpha})-\bm{y}_{\alpha})\le 0, \nonumber \\
(\hat{\bm{\beta}}-\bm{\beta})^{T}(\text{Proj}(\hat{\bm{\beta}},\bm{y}_{\beta})-\bm{y}_{\beta})\le 0.
\end{align}
From \eqref{adap_law} and \eqref{proj_property}, we can imply that
\begin{align}
\label{proj imply}
\tilde{\bm{\alpha}}^{T}\bm{\Gamma}^{-1}\dot{\tilde{\bm{\alpha}}} \le \tilde{\bm{\alpha}}^{T}\bm{y}_{\alpha}-\tilde{\bm{\alpha}}^{T}\bm{\Gamma}^{-1}\dot{\bm{\alpha}}, \nonumber \\ 
\tilde{\bm{\beta}}^{T}\bm{\Gamma}^{-1}\dot{\tilde{\bm{\beta}}} \le \tilde{\bm{\beta}}^{T}\bm{y}_{\beta}-\tilde{\bm{\beta}}^{T}\bm{\Gamma}^{-1}\dot{\bm{\beta}}.
\end{align}

We now replace \eqref{CLF eta parts}, \eqref{RES eta tilde} and \eqref{proj imply} to \eqref{dotV}, which results in
\begin{align}
\dot{\tilde{V}} & \le
-\lambda\tilde{\bm{\eta}}^{T}\bm{P}\tilde{\bm{\eta}} + \epsilon_{\tilde{V}} \nonumber \\
&+
\tilde{\bm{\alpha}}^{T}(\bm{y}_{\alpha}+\bm{H}^{T}\bm{P}\tilde{\bm{\eta}}||\bm{\eta}||)-\tilde{\bm{\alpha}}^{T}\bm{\Gamma}^{-1}\dot{\bm{\alpha}} \nonumber \\
&+
(\bm{y}_{\alpha}^{T}+\tilde{\bm{\eta}}^{T}\bm{P}\bm{H}||\bm{\eta}||)\tilde{\bm{\alpha}}-\dot{\bm{\alpha}}^{T}\bm{\Gamma}^{-1}\bm{\alpha}
\nonumber \\
&+
\tilde{\bm{\beta}}^{T}(\bm{y}_{\beta}+\bm{H}^{T}\bm{P}\tilde{\bm{\eta}})-\tilde{\bm{\beta}}^{T}\bm{\Gamma}^{-1}\dot{\bm{\beta}} \nonumber \\
&+
(\bm{y}_{\beta}^{T}+\tilde{\bm{\eta}}^{T}\bm{P}\bm{H})\tilde{\bm{\beta}}-\dot{\bm{\beta}}^{T}\bm{\Gamma}^{-1}\tilde{\bm{\beta}}
\end{align}

So, by using the chosen projection functions \eqref{eq:proj_fun}, then we conclude that.
\begin{align}
\label{RES_CLF bound}
\dot{\tilde{V}}+\lambda\tilde{V} \le  \epsilon_{\tilde{V}} +
\lambda\tilde{\bm{\alpha}}^{T}\bm{\Gamma}^{-1}\tilde{\bm{\alpha}}+
\lambda\tilde{\bm{\beta}}^{T}\bm{\Gamma}^{-1}\tilde{\bm{\beta}} \nonumber \\
-\tilde{\bm{\alpha}}^{T}\bm{\Gamma}^{-1}\dot{\bm{\alpha}} -\dot{\bm{\alpha}}^{T}\bm{\Gamma}^{-1}\tilde{\bm{\alpha}}
\nonumber \nonumber \\
-\tilde{\bm{\beta}}^{T}\bm{\Gamma}^{-1}\dot{\bm{\beta}} -\dot{\bm{\beta}}^{T}\bm{\Gamma}^{-1}\tilde{\bm{\beta}}.
\end{align}

We assume that the uncertainties $\bm{\alpha}$, $\bm{\beta}$ and their time derivatives are bounded. Furthermore, the projection operators \eqref{adap_law} will also keep $\tilde{\bm{\alpha}}$ and $\tilde{\bm{\beta}}$ bounded (see \cite{L1:UnkownNonlinearities:ACC08} for a detailed proof about these properties.) We define these bounds as follows:
\begin{align}
\label{bounded conditions for CLF proof}
||\tilde{\bm{\alpha}}|| \le& \tilde{\bm{\alpha}}_b ,~~
||\tilde{\bm{\beta}}|| \le \tilde{\bm{\beta}}_b , \nonumber \\
||\dot{\bm{\alpha}}|| \le& \dot{\bm{\alpha}}_b ,~~
||\dot{\bm{\beta}}|| \le \dot{\bm{\beta}}_b. 
\end{align}
Combining this with \eqref{RES_CLF bound}, we have,
\begin{align}
\label{relaxed CLF condition deltaV}
&\dot{\tilde{V}}+\lambda\tilde{V} \le \lambda\delta_{\tilde{V}}, 
\end{align}
where 
\begin{align}
\delta_{\tilde{V}}=2||\bm{\Gamma}||^{-1}(\tilde{\bm{\alpha}}_{b}^2+\tilde{\bm{\beta}}_{b}^{2}+\frac{1}{\lambda}\tilde{\bm{\alpha}}_{b}\dot{\bm{\alpha}}_{b}+\frac{1}{\lambda}\tilde{\bm{\beta}}_{b}\dot{\bm{\beta}}_{b}) + \frac{1}{\lambda}\epsilon_{\tilde{V}}.
\end{align}
Thus, if $\tilde{V} \ge \delta_{\tilde{V}}$ then $\dot{\tilde{V}}\le 0$. As a result, we always have $\tilde{V} \le \delta_{\tilde{V}}$. 
In other words, by choosing the adaptation gain $\bm{\Gamma}$ sufficiently large and $\bm{P}$ quite small, we can limit the Control Lyapunov Function \eqref{CLF tilde} in an arbitrarily small neighborhood $\delta_{\tilde{V}}$ of the origin. 
According to \eqref{Amatrix} and \eqref{eq:Lyp}, achieving a small value for $\bm{P}$ depends on choosing a proper value for $\bm{K}_P$, $\bm{K}_D$, and $\bm{Q}$. Therefore, the value of PD gains affects the stability of the whole system.
Finally, the tracking errors between the dynamics model \eqref{EtaClosedLoopUncertainty} and the reference model \eqref{ref_model}, $\tilde{\bm{\eta}}$, and the error between the real and estimated uncertainty, $\tilde{\bm{\alpha}}$, $\tilde{\bm{\beta}}$ are bounded as follows:
\begin{align}
||\tilde{\bm{\eta}}|| \le \sqrt{\frac{\delta_{\tilde{V}}}{||\bm{P}||}} ,
||\tilde{\bm{\alpha}}|| \le \sqrt{||\bm{\Gamma}||\delta_{\tilde{V}}} ,||\tilde{\bm{\beta}}|| \le \sqrt{||\bm{\Gamma}||\delta_{\tilde{V}}}.
\end{align}
\section{Numerical Validation}
\label{sec:simulation}

\begin{figure}[b!]
	\center
	\includegraphics[width=0.7\linewidth]{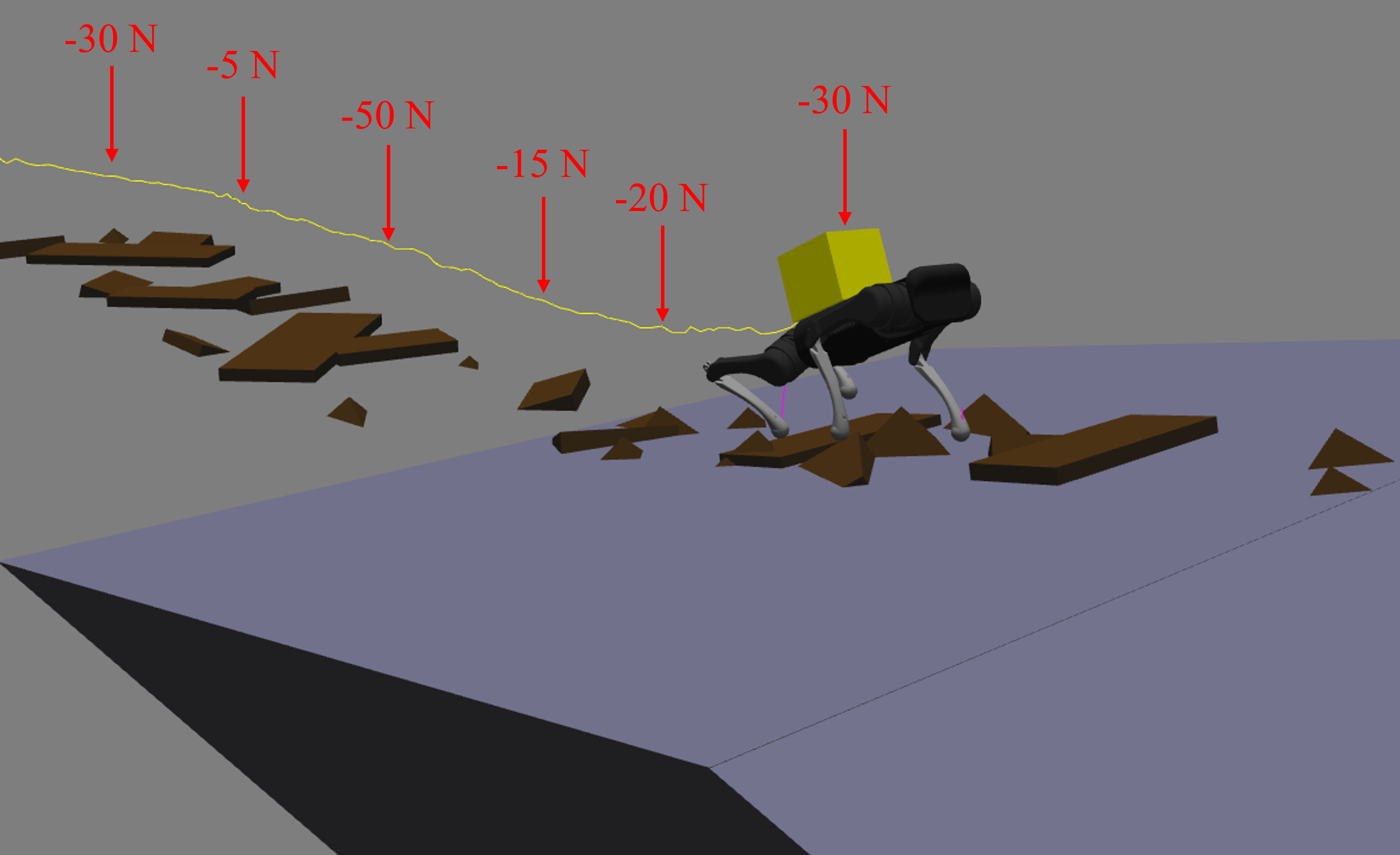}
	\caption{{\bfseries The A1 Robot Simulation.}  The robot walking on high-sloped terrain with obstacles while carrying a 6kg load with varying external force. Simulation video: \protect\url{https://youtu.be/UHz8h-CuZ6w}} 
	\label{fig:uneven terrain simulation}
	\vspace{-1.5em}
\end{figure}

This section presents validation conducted in a high-fidelity simulation of the A1 robot. The control system is implemented in ROS Melodic with Gazebo 9 simulator. For the adaptive controller, we set the parameters as presented in table \ref{tab:parameter}. We set the adaptive gains larger for height, pitch, and roll because our simulation emphasizes the adaptation for these 3 states. The robot is simulated to carry a load of up to 50\% of the robot weight during walking based on the approach we have developed in previous sections. \figref{fig:uneven terrain simulation} shows that our proposed adaptive force control can enable the robot to climb stably to an uneven slope while adapting to time-varying model uncertainty (carrying an unknown load while being pushed by a varying external force in the z-direction). This result not only proves our proposed approach's capability in adapting to a high level of model uncertainty but also illustrates that our approach can retain the robustness of force control in navigating rough terrains.  


\begin{table}[bt!]
	\centering
	\caption{Controller Setting}
	\label{tab:parameter}
	\begin{tabular}{|c|c|c|c|}
		\hline
		Parameter & Value & Parameter & Value\\
		\hline
		$\bm{K}_P$  & $diag(30, 30, 50, 80, 80, 80)$  & $\gamma_1$ & 0.01 \\[.5ex]
		$\bm{K}_D$  & $diag(10, 10, 10, 50, 50, 50)$ & $\gamma_2$ & 0.001 \\[.5ex]
		$\bm{S}$ & $diag(5, 5, 10, 50, 25, 20)$ & $\zeta$ & 0.7 \\[.5ex]
		$\bm{\Gamma}$ & $diag(1, 1, 5, 2, 5, 1) \times 10^3$ & $\omega_n$ & 400\\[.5ex]
		\hline
	\end{tabular}
\end{table}

\begin{figure}[hbt!]
	\centering
	\subfloat[Non-adaptive control]{\includegraphics[width=0.47\linewidth] {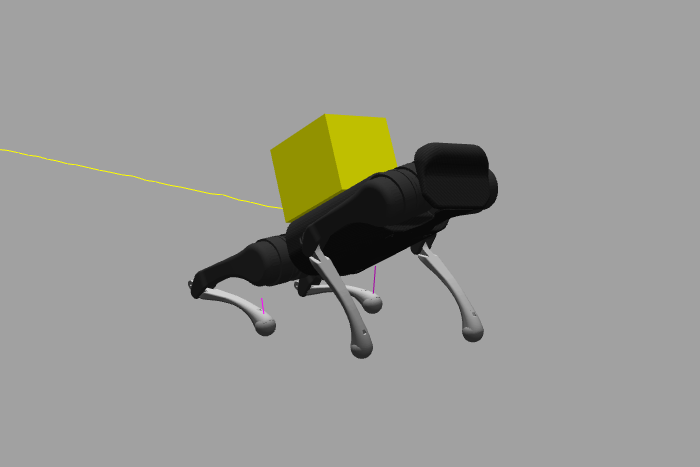}}
	\hfill
	\subfloat[Adaptive control]{\includegraphics[width=0.47\linewidth] {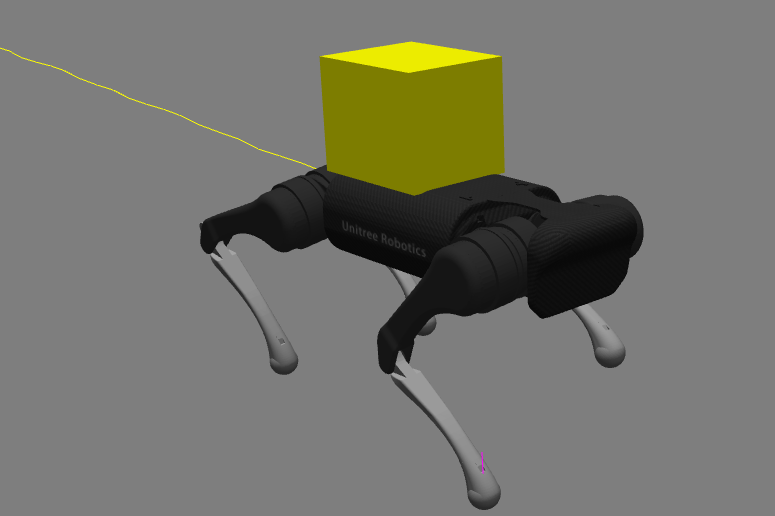}}
	\hfill
	\vspace{-1.0em}
	\subfloat{\includegraphics[width=0.5\linewidth] {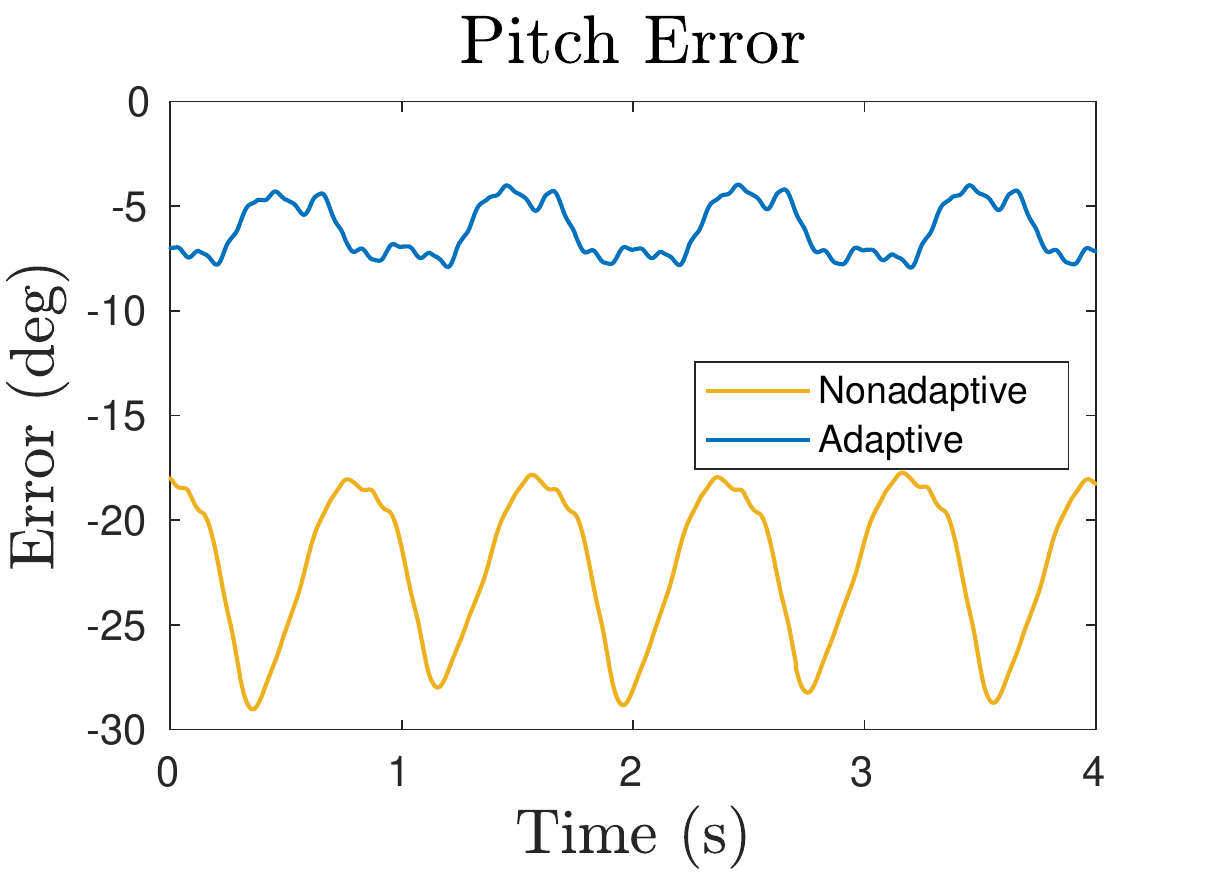}}
	\hfill
	\subfloat{\includegraphics[width=0.5\linewidth] {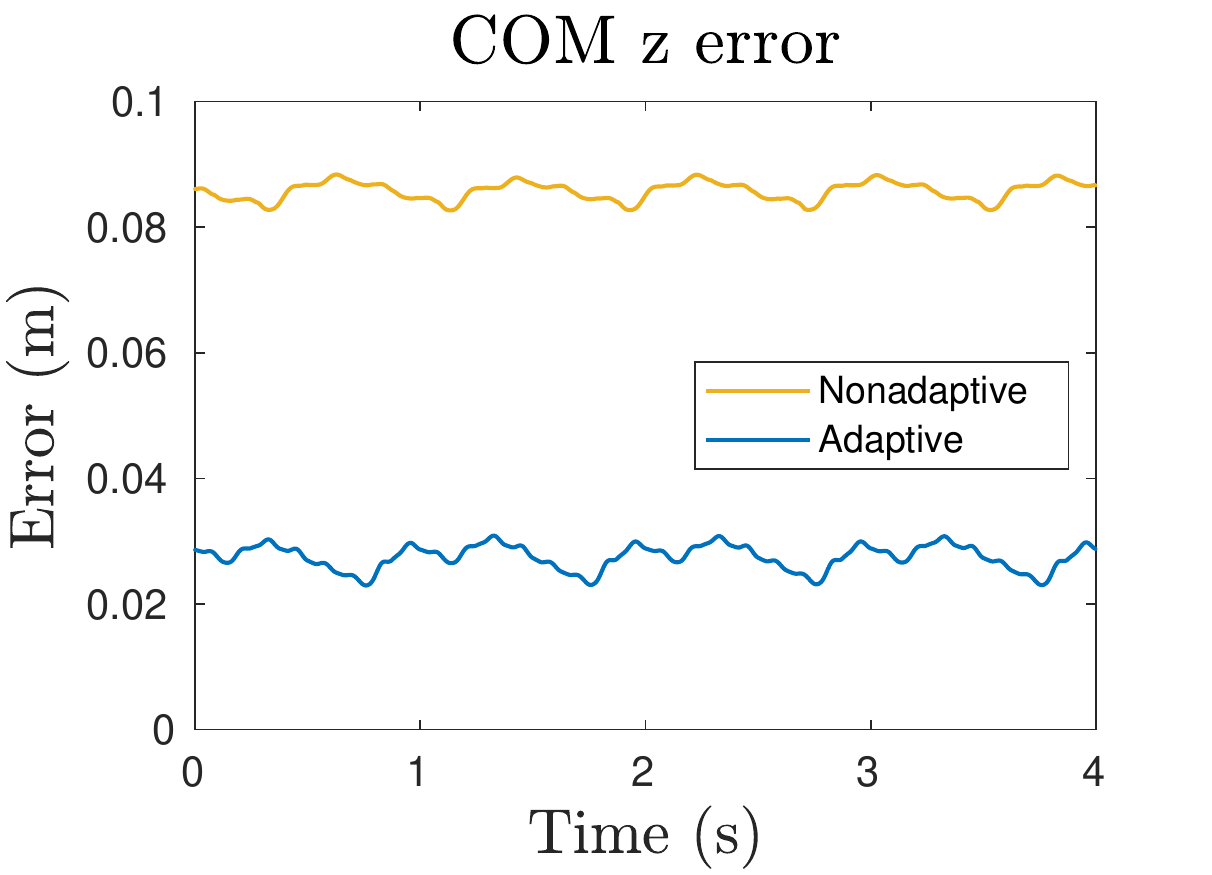}}
	
	\caption{Simulation results of the robot walking with a 6-kg load on the back using (a) non-adaptive control and (b) adaptive control.}
	\label{fig:simulation result}
	\vspace{-1.5em}
\end{figure}

In the simulation, to introduce model uncertainty in both mass and inertia to the robot dynamics, a 6-kg load, which is 50\% of the robot's weight, is applied to the robot's back. As shown in \figref{fig:simulation result}, while the non-adaptive controller fails to keep the robot balanced with about $30^o$ error in the pitch angle and $9~cm$ error in the robot height, our proposed adaptive control can keep the robot balanced with a very small tracking error in both pitch angle (less than $8^o$) and robot height (less than $3~cm$). Since our simulation does not emphasize the significant difference in tracking errors of position along the x-axis, y-axis, roll, and yaw angles, we select to show plots of tracking errors in the z-axis and the pitch angle.

%

\section{Experimental Validation}
\label{sec:experiment}
\begin{figure}[t!]
	\centering
	\subfloat[Non-adaptive control]{\includegraphics[width=0.5\linewidth]{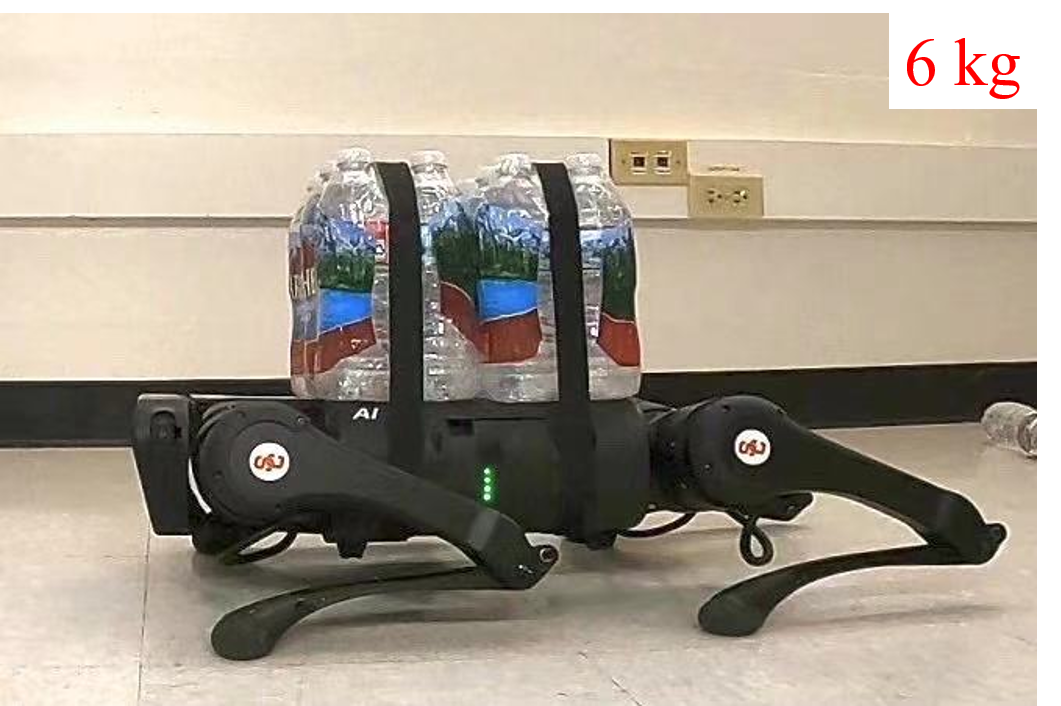}}
	\hfill
	\subfloat[Adaptive control]{\includegraphics[width=0.5\linewidth]{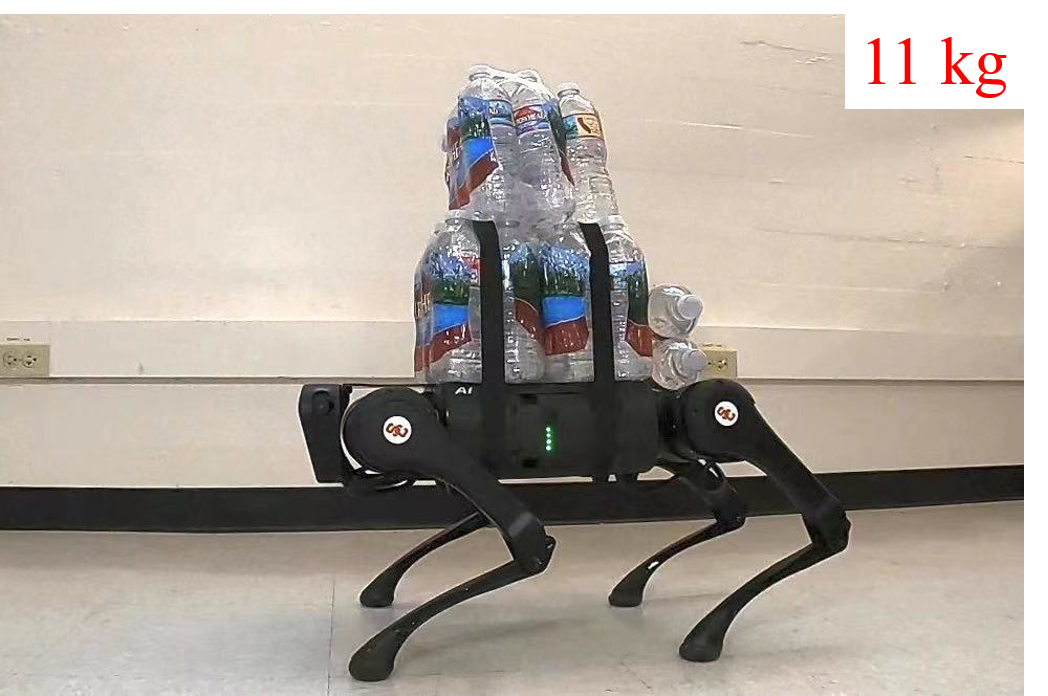}}
	\hfill
	\subfloat{\includegraphics[width=0.5\linewidth] {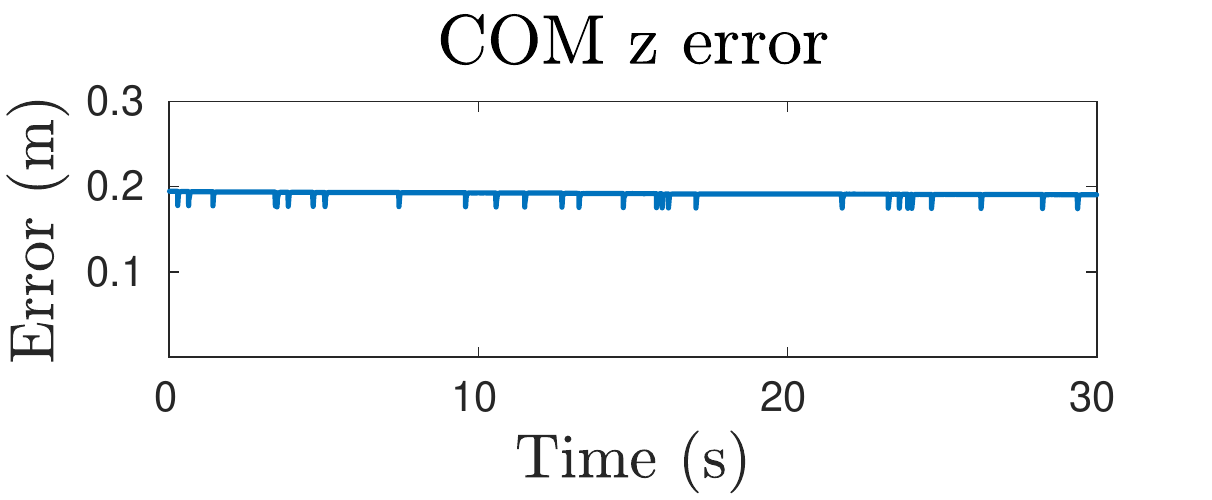}}
	\hfill
	\subfloat{\includegraphics[width=0.5\linewidth]
	{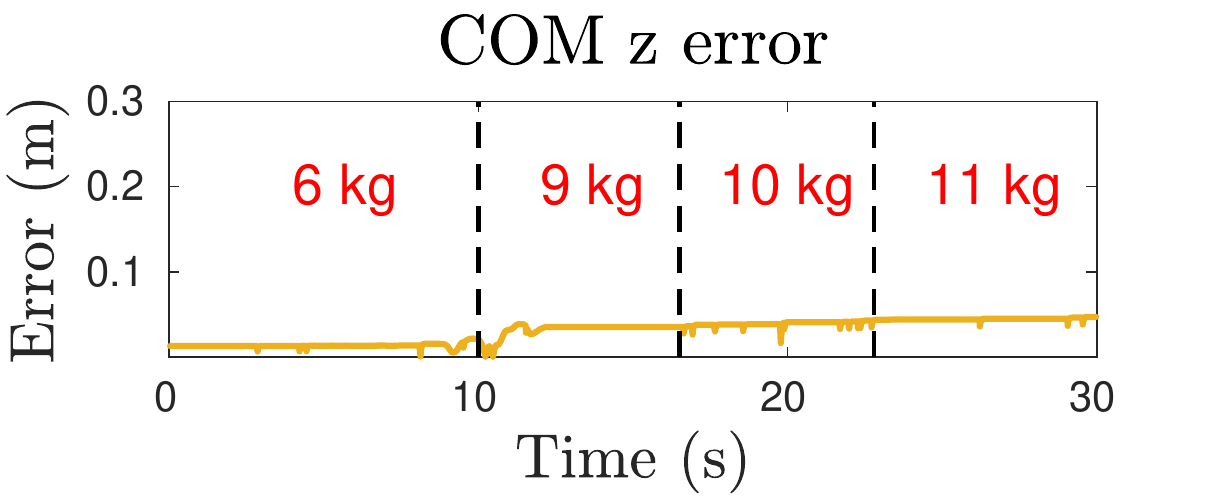}}
	\caption{{\bfseries Standing experiment results.} a) Non-adaptive controller with 6-kg load; b) Adaptive controller with up to 11-kg load.}
	\label{fig:z_varying_load}
	\vspace{-1.8em}
\end{figure}

We have also successfully implemented our approach on the real robot hardware (see \figref{fig:uneven terrain}). This section will show our comparison between (a) non-adaptive control and (b) adaptive control for the problem of quadruped robots standing and walking while carrying an unknown load on the trunk. In the experiment, the weight and inertia of the load are unknown to neither the non-adaptive control nor the adaptive control approach.
To demonstrate the effectiveness of our approach, we tested the baseline controller and adaptive controller with the same load and control parameters. We added the load to the robot throughout the experiment until it fails or meets the expectation from our simulation. For standing, we added loads gradually from $1~kg$ for the robot to stand up to a height of $30cm$. As presented in \figref{fig:z_varying_load}, with the baseline non-adaptive controller, the robot could barely stand up when the load is added to $6kg$, resulting in a large tracking error of in z-direction of approximately $20~cm$. With the adaptive controller, the robot can stand up with $6kg$ load on its back with a small tracking error (about $2~cm$), plus it can handle a time-varying load of up to $11~kg$. With the adaptive control, the tracking error in the z-axis is still less than $5~cm$ even with an unknown 11-kg load applied to the robot.



\begin{figure}[bt!]
	\centering
	\subfloat[Non-adaptive control]{\includegraphics[width=0.48\linewidth]
	{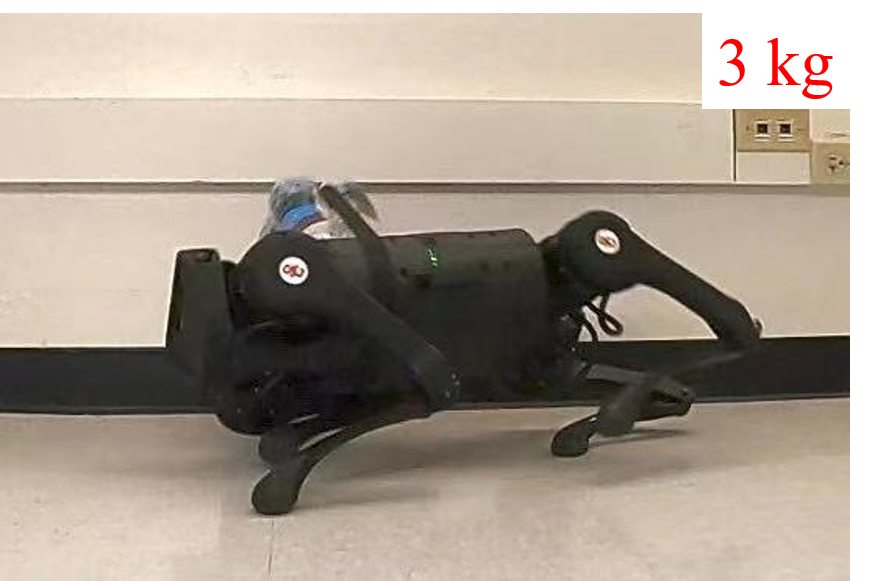}}
	\hfill
	\subfloat[Adaptive control]{\includegraphics[width=0.48\linewidth]
	{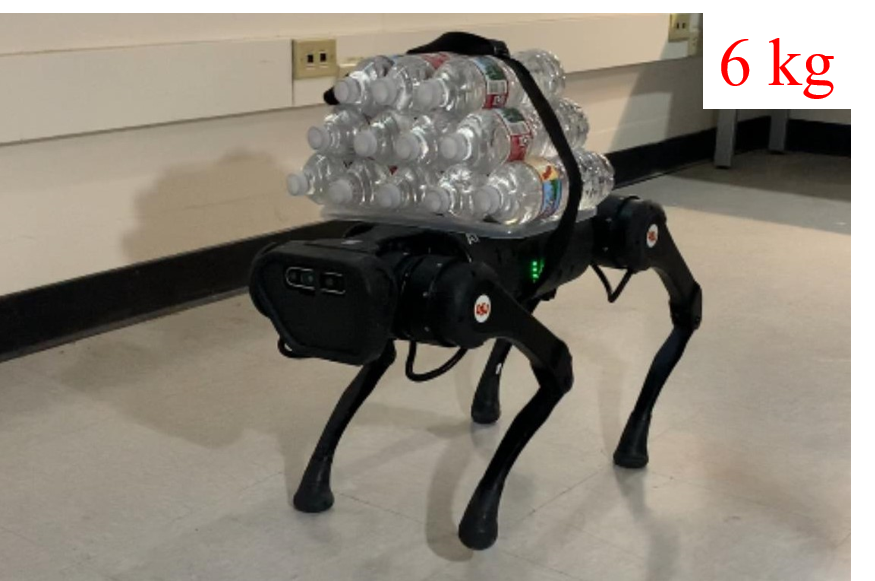}}
	\hfill
	\vspace{-1.0em}
	\centering
	\subfloat{\includegraphics[width=0.5\linewidth] {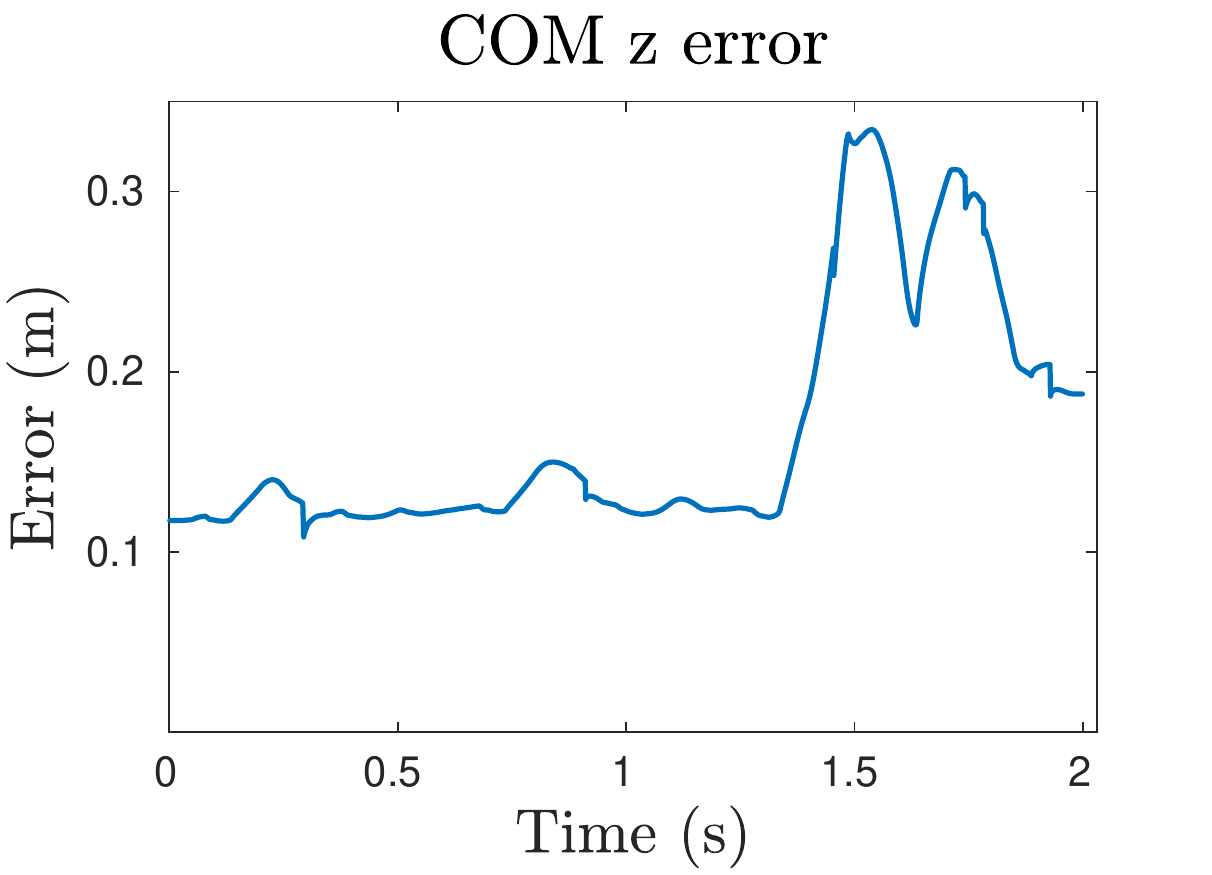}}
	\hfill
	\subfloat{\includegraphics[width=0.5\linewidth] {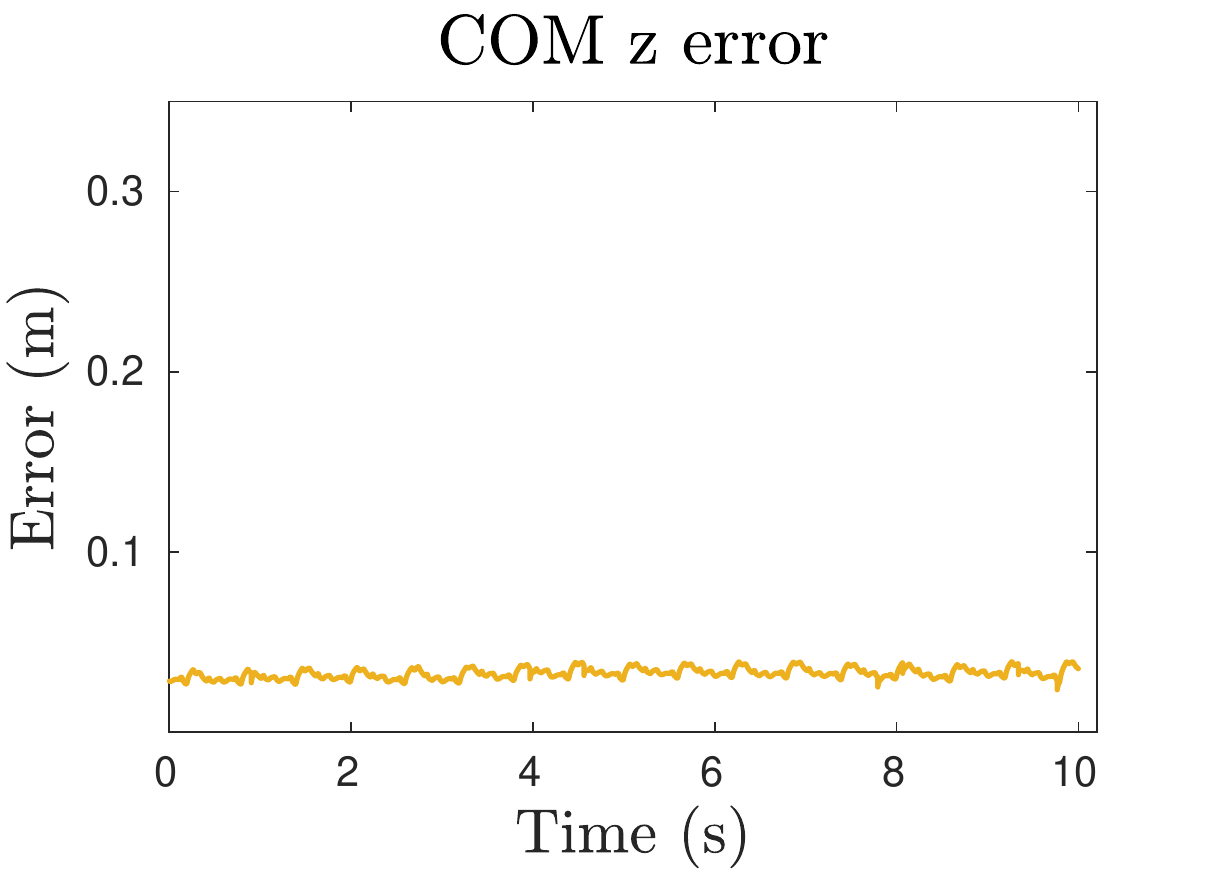}}
	\caption{{\bfseries Walking experiment results.} a) Non-adaptive controller walking with 3-kg load; b) Adaptive controller walking with a 6-kg load. }
	\label{fig:z_error_walk}
	\vspace{-1.5em}
\end{figure}

\begin{figure}[t!]
	\centering
	\subfloat{\includegraphics[width=0.5\linewidth] {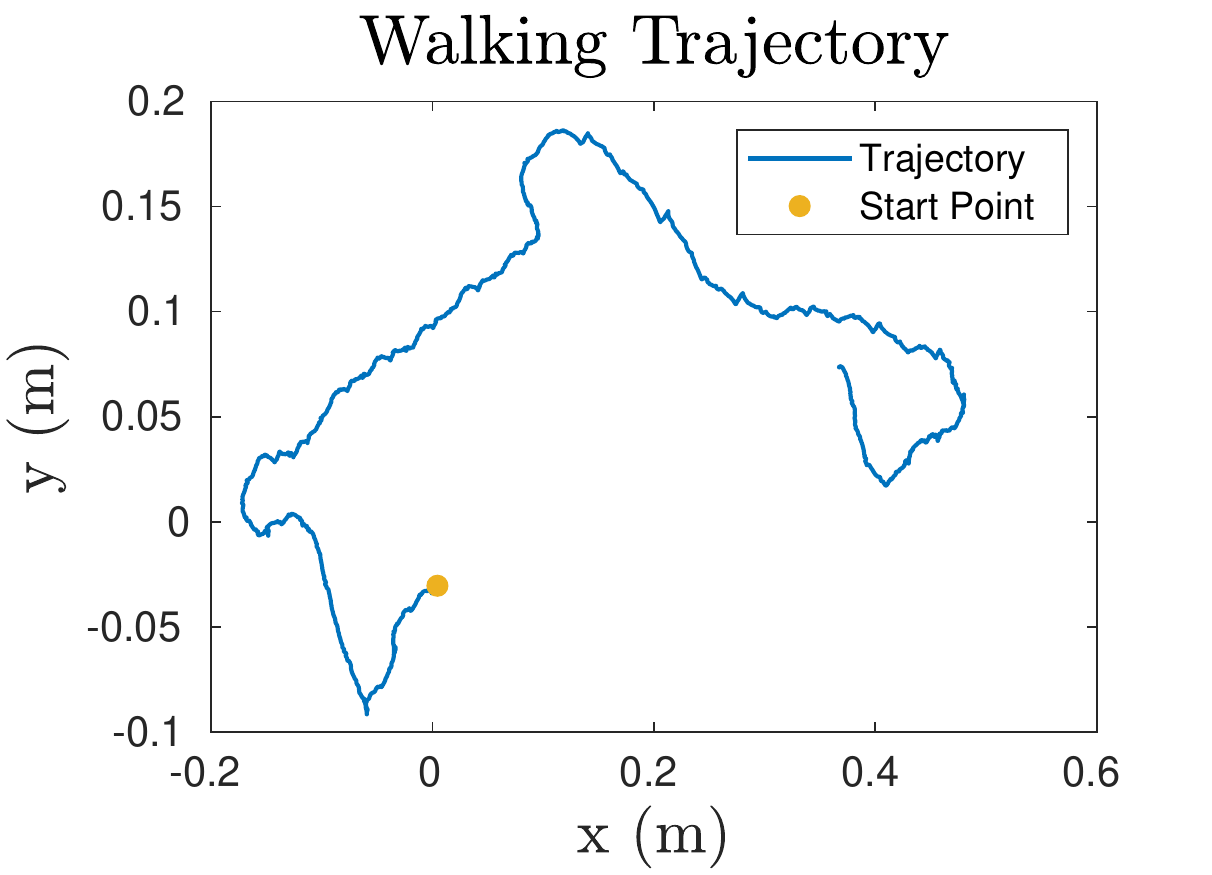}}
	\hfill
	\subfloat{\includegraphics[width=0.5\linewidth] {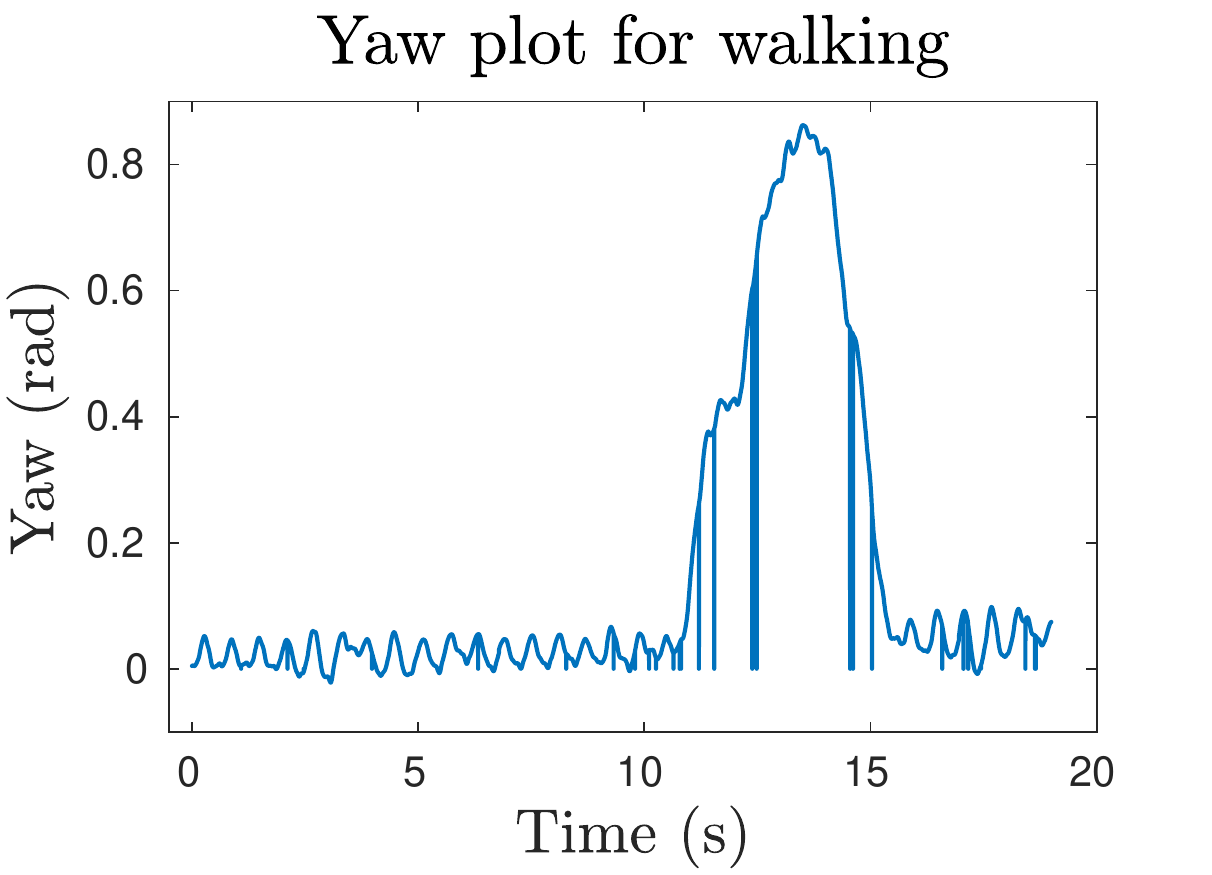}}
	\hfill
	
	\caption{{\bfseries Walking experiment plot} a) robot trajectory plot; b) yaw plot for walking.}
	\label{fig:walking_snapshot}
	\vspace{-1.5em}
\end{figure}

In the walking experiment, as presented in \figref{fig:z_error_walk}, the robot fails with the non-adaptive controller when a load of only 3kg is applied to the robot. The robot cannot keep balance due to model uncertainty and failed to the right within 2 seconds of walking. In comparison, our proposed adaptive force control can allow the robot to walk stably in all directions while carrying an unknown load of 6 kg, which is 50\% of the robot's weight. \figref{fig:walking_snapshot}, shows the path of the robot walking forward, backward, side-way, and turning while carrying the heavy load. The spikes shown in \figref{fig:z_varying_load} and \figref{fig:walking_snapshot} result from the noise of the IMU during the experiment. More details of the walking experiment are presented in the supplemental video. Although there is a small constant error for the adaptive controller, it does not contradict the algorithm we have developed because the controller system we have designed guarantees input-to-state stability. The experiment results have clearly demonstrated the advancements of our proposed approach.



\section{Conclusion}
\label{sec:conclusion}
In summary, we have presented a novel control system that incorporates adaptive control into force control for dynamic legged robots walking under uncertainties. 
We have demonstrated the effectiveness of our proposed approach using both numerical and experimental validations. In simulation, we have shown that our proposed approach can allow the robot to climb up an uneven slope while carrying an unknown load of up to 6 kg (50\% of the robot weight). In experiment, we have successfully implemented our proposed adaptive force control for quadruped robots standing and walking while carrying an unknown heavy load on the trunk. The experiment has presented impressive results when the robot can carry up to 11 kg of unknown load (92\% of the robot weight) while standing with 4 legs while maintaining a tracking error of less than $5~cm$ in the robot height. In addition, with the adaptive controller, the robot can also walk stably in all directions with 6 kg of load on its trunk. In contrast, the baseline non-adaptive controller fails within 2 seconds of walking with only 3 kg of load. 
Our work has shown that the proposed adaptive force control not only can adapt to large model uncertainty but can also leverage the advantages of force control in navigating rough terrains for legged robots. 
In the future, we will extend our framework to achieve more dynamic gaits.

\balance

\bibliographystyle{ieeetranS}
\bibliography{bibdata,biped,Ames_ref,clf_ref,robust_ref,L1_ref,ICRA2019,ACC2020}

\end{document}